\documentclass[10pt,twocolumn,letterpaper]{article}

\usepackage{iccv}
\usepackage{times}
\usepackage{epsfig}
\usepackage{graphicx}
\usepackage{amsmath}
\usepackage{amssymb}
\usepackage{float}
\usepackage{multirow}
\usepackage{subcaption}
\usepackage{placeins}

\usepackage[pagebackref=true,breaklinks=true,colorlinks,bookmarks=false]{hyperref}

\newcommand{\model}{Interaction-CLIP }
\newcommand{\nickname}{iCLIP}

\iccvfinalcopy 

\def\iccvPaperID{8} 

\ificcvfinal\fi
\begin{document}

\title{Interaction-Aware Prompting for Zero-Shot Spatio-Temporal Action Detection}

\author{Wei-Jhe Huang$^1$ \quad
Jheng-Hsien Yeh$^1$ \quad
Min-Hung Chen$^2$ \quad
Gueter Josmy Faure$^3$ \quad
Shang-Hong Lai$^1$ \\
$^1$National Tsing Hua University, Taiwan \quad
$^2$NVIDIA \quad
$^3$National Taiwan University
\\
{\tt\small  
\{weijhie, goodcharlie1018, vitec6, josmyfaure\}@gmail.com \quad lai@cs.nthu.edu.tw}
}

\twocolumn[{%
\renewcommand\twocolumn[1][]{#1}%
\maketitle
\begin{center}
    \centering
    \captionsetup{type=figure}
    \includegraphics[width=1.0\textwidth]{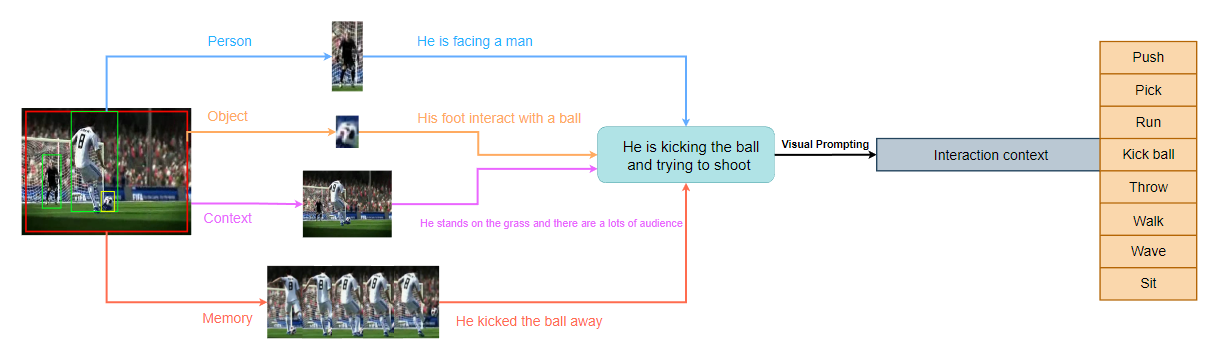}
\caption{\textbf{Interaction context for Prompting.}
Our method aims to capture various types of information related to a person's interaction with their environment. \textbf{(1)} Person-person information contributes to their interaction.  \textbf{(2)} Person-object information contributes what is the person doing with the object. \textbf{(3)} Person-context gives the environment information.  \textbf{(4)} Person-memory reference before and after frames to conclude what happened. By combining the interaction token with text tokens, we create interaction-aware prompting that enhances the model's ability to generalize to unseen action classes.
}
\label{visualization}
\end{center}
}]


\maketitle
\ificcvfinal\thispagestyle{empty}\fi

\label{sec:Abstract}
\begin{abstract}
\vspace{-5pt}
The goal of spatial-temporal action detection is to determine the time and place where each person's action occurs in a video and classify the corresponding action category. Most of the existing methods adopt fully-supervised learning, which requires a large amount of training data, making it very difficult to achieve zero-shot learning. In this paper, we propose to utilize a pre-trained visual-language model to extract the representative image and text features, and model the relationship between these features through different interaction modules to obtain the interaction feature. In addition, we use this feature to prompt each label to obtain more appropriate text feature. Finally, we calculate the similarity between the interaction feature and the text feature for each label to determine the action category. Our experiments on J-HMDB and UCF101-24 datasets demonstrate that the proposed interaction module and prompting make the visual-language features  better aligned, thus achieving excellent accuracy for zero-shot spatio-temporal action detection. The code
will be available at \href{https://github.com/webber2933/iCLIP}{https://github.com/webber2933/iCLIP}.
\end{abstract}
\section{Introduction}
\label{sec:intro}
Video understanding refers to the process of extracting meaningful information from video data, which includes tracking objects and people, recognizing actions and events, and understanding the overall context and narrative of a video. Two important tasks within video understanding are action recognition and action detection. Action recognition aims to classify a whole video into an appropriate class, without specifying the temporal locations. For example, a video may be classified as "walking" or "running" based on the actions performed by the people in the video. As for action detection, in addition to recognizing specific actions or activities, it also needs to specify the temporal duration (i.e., start/end time) of the action in an untrimmed video. Compared with these two tasks, spatio-temporal action detection is more challenging because it requires localizing the space and time where each person’s action occurs, and classifying the associated action class at the same time.

Conventional action detection methods rely on supervised learning and require a large number of videos per class with costly annotation for training. However, this approach severely limits the scalability of the task. To address this problem, few-shot, and zero-shot learning have attracted much attention. Few-shot learning aims to recognize new classes with only a few examples, whereas zero-shot learning aims to predict unseen classes, which are different from the classes in the training data. In this scenario, the commonly used approach is to project training class names into some semantic space. When the semantic space is aligned with a visual feature space, a model trained from existing classes can be applied to the new ones. This approach can greatly improve the scalability of action detection to a larger number of classes without requiring costly annotations for training.

Recently, large pretrained Visual-Language models \cite{radford2021learning, jia2021scaling, yuan2021florence} are often used in advancing zero-shot learning, because these models offer a strong alignment between text and visual modality by projecting their embedding into the same feature space. With a large amount of training data, these models have a strong ability for generalization and therefore perform well in many different zero-shot tasks like image classification. Due to its good performance on recognizing images, there are recent works \cite{ju2022prompting, nag2022zero} combining with these models to achieve zero-shot action detection, whilst existing works mostly focus on temporal detection, which does not localize in the spatial domain for action detection.

In this paper, we explore spatio-temporal action detection
in the zero-shot scenario (ZSSTAD). The task involves detecting all persons in a video and identifying their action classes, even when the actions are unseen or unknown. While some prior works \cite{Jain_2015_ICCV,Mettes_2017_ICCV,mettes2021object,mettes2022universal} simply exploit pre-trained models to make an inference with all labels of the dataset, we aim to have better performance in
ZSSTAD by training additional parameters. To tackle this challenging problem, we propose a new architecture called \model (\nickname), which uses CLIP \cite{radford2021learning} encoders to extract important feature, including person, object, and context, for recognizing actions. We develop interaction modules to combine the information that people interact with these features. Through this approach, we obtain an "interaction feature" that can more accurately describe the action. Besides, We add a module called Interaction-Aware Prompting, which uses our interaction feature to prompt the original CLIP text embedding of each label. This helps us describe each label more completely through the action of the current person, thus increasing its discriminability and allowing us to better classify the action.
In summary, our contributions are summarized as follows:
\begin{itemize}
    \item [-]
    We explore the zero-shot spatial-temporal action detection problem, which has not been studied much before. Our method can learn how to model a person's actions through the training class and transfer knowledge to the unseen class.
    \item [-]
    To solve this problem, we introduce a novel method, \nickname, that integrates human interaction features and learning-to-prompt in the same framework.
    \item [-]
    The experiments show that our interaction feature can be used to detect unseen actions more accurately than the naive CLIP image feature. This indicates that the proposed model can make the pretrained visual-language features (e.g., CLIP) better aligned in the shared embedding space.
\end{itemize}


\begin{figure*}[htbp]
\begin{center}
   \includegraphics[scale=0.6]{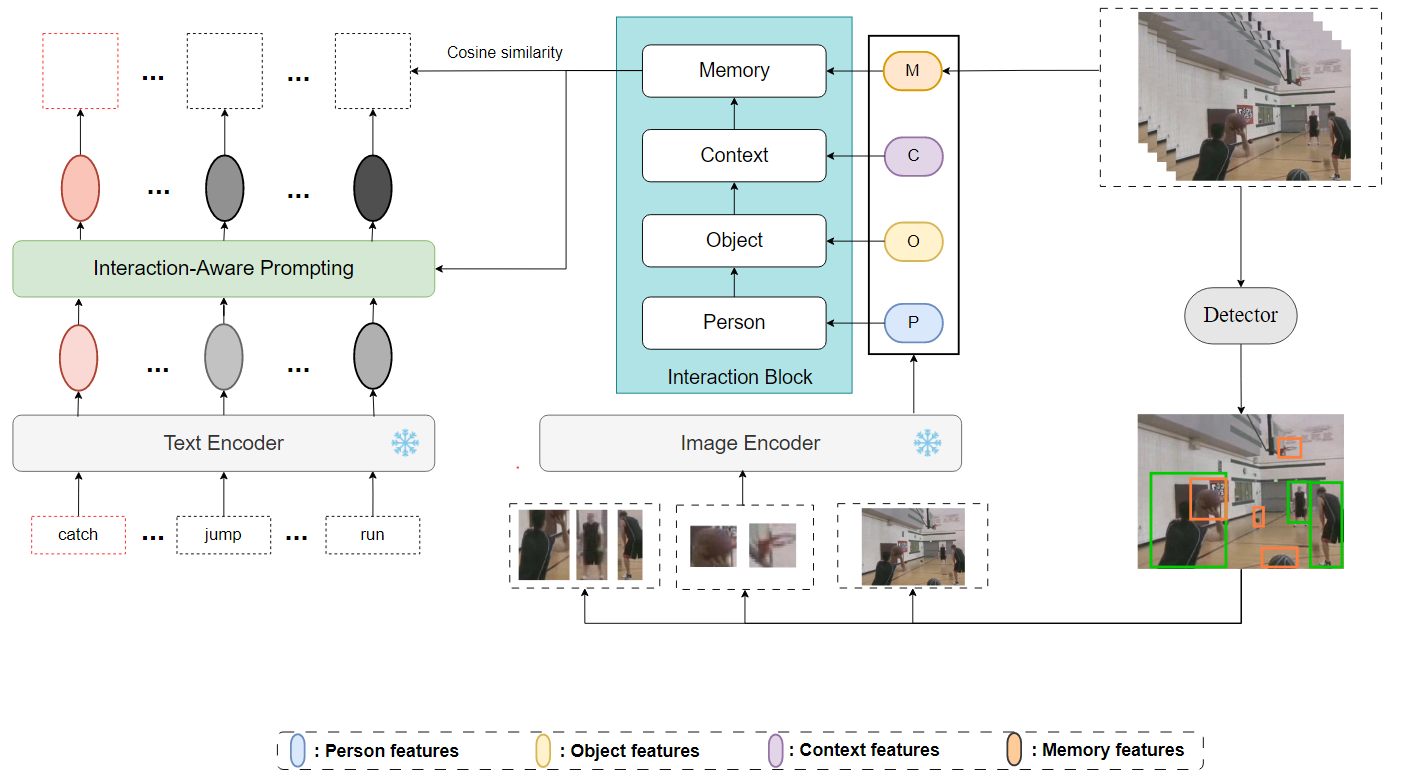}
\end{center}
   \caption{\textbf{Overview of \nickname \space network:}
  \textbf{(1)} The left shows that we use a text encoder to compute text embedding of each action label. \textbf{(2)} The right figure shows that a frame is cropped into person images, object images, and a context image. Then we use an image encoder to compute their features. \textbf{(3)} The middle shows that these features serve as  input into interaction blocks, which can aggregate the information of different interaction, and give us an "interaction feature". Finally, we use Interaction-Aware Prompting to combine the text embedding with the interaction feature and calculate the similarity between visual and text embedding.  
   }
\label{fig:overview}

\end{figure*}
\section{Related work}
\label{Related}
\subsection{Vision-Language for Video Understanding}
The detection of actions is often expressed through language, either as individual words or complete sentences (e.g., "run," "drive a car"). As such, vision-language models offer a natural approach to action detection. The majority of research in this field has utilized the CLIP model \cite{radford2021learning}, which is highly effective at matching images with their corresponding verbal descriptions. One recent study \cite{nag2022zero} proposed a framework that uses CLIP as a backbone and an alignment method for image-video adaptation, enabling parallel localization and classification of actions. Another approach that emphasizes efficiency is the use of prompting language models pre-trained on image-text pairs, which has shown success in downstream video understanding tasks, including action detection \cite{ju2022prompting}. Despite the progress made in this area, there is currently no research exploring the use of vision-language models for spatio-temporal action detection in the literature.

\subsection{Spatio-Temporal Action Detection}
The detection of actions in both space and time is a crucial area of research, with broad applications spanning from video understanding to autonomous vehicles. There exists a rich body of work in this field, with numerous studies published \cite{feichtenhofer2019slowfast, girdhar2019video, li2019collaborative, pan2021actor, tang2020asynchronous, Faure_2023_WACV}. The current standard approach involves utilizing a 3D CNN backbone to extract video features, followed by cropping the region of interest (ROI) - typically the person to be classified, along with any relevant surrounding objects. Recent research in spatio-temporal action detection emphasized modeling interactions between the person being classified and the environment \cite{pan2021actor, tang2020asynchronous, wu2019long, Faure_2023_WACV, materzynska2020something, tang2020asynchronous, yang2021beyond, zhou2021graph}. To extract meaning from these diverse feature sets, including person-object and person-context interactions, most studies rely on cross-attention. Despite the abundance of research on zero-shot temporal action detection, the topic of zero-shot spatio-temporal action detection has not been explored much at the moment.
\subsection{Zero-shot Action Detection}
Zero-shot temporal action detection (ZSTAD) has been a well-explored topic in recent years, primarily due to the emergence of vision-language models (ViL), with the CLIP \cite{radford2021learning} model being the main backbone for this task. Previous works \cite{lin2017single, gan2015exploring, mettes2017spatial} utilized word embedding methods such as GLoVE \cite{qin2017zero} or word2vec for ZSTAD. More recent studies, such as \cite{nag2022zero, ju2022prompting, ni2022expanding, chen2021elaborative},  employed CLIP to extract features from text data, specifically labels, and leveraged prompting to align visual and text features to achieve zero-shot detection. Notably, despite the better image-text alignment capability for CLIP, previous works have utilized other feature extraction methods, such as \cite{mandal2019out}, which employed a GAN approach to generate "fake" text features for previously unseen action categories. Given the recent success of CLIP, we adopt it as our feature extractor, as it is pre-trained on image-text pairs and generates superior cross-modal representations. However, our approach is fundamentally distinct from the previous works as we tackle the more complex task of zero-shot spatio-temporal action detection (ZSSTAD).

\section{Proposed Method}
\label{Method}

\subsection{Pretrained Visual-Language Model}
In this paper, we use CLIP~\cite{radford2021learning} as our pretrained visual-language model because it provides an embedding space that has rich visual and semantic information, allowing us to align image and text to the same space. It mainly consists of an image encoder and text encoder, and uses a large number of image-text pairs crawled on the Internet for model training. In the process of training, it takes N corresponding image-text pairs each time and obtains the embedding of each image and text through the encoder, then calculates the cosine similarity between these $N\times N$ combinations. Through cross-entropy loss, that is, maximizing the cosine similarity of the correct N image-text pairs, and simultaneously minimizing the cosine similarity of other wrong pairs, it can jointly optimize image and text encoders. Because CLIP has been quite successful for zero-shot image classification tasks, we choose to make further improvements based on its feature embedding, so that we can learn more appropriate features to detect unseen actions.

\subsection{Overall Framework}
The overall structure of the proposed model, as shown in Figure \ref{fig:overview}, is a multimodal framework composed of two modalities, visual and text. Each of them is responsible for modeling the interaction feature of the person and the text feature of each label, respectively. Both the CLIP image encoder and text encoder are frozen during our model training. Lastly, we combine the two parts by calculating the cosine similarity between the interaction feature and each label feature. 

The visual component of our approach involves isolating the frame containing the target person of interest, and subsequently extracting relevant portions of the image which is helpful to the classification of actions. These portions include persons, as well as objects overlapping with any person in the scene. Following this, we employ a pre-trained image encoder from CLIP to obtain person-specific, object-specific, and contextual features from both the extracted portions and the entire frame. To model the relationship between the target individual and their surrounding environment, similar to Faure et al. (2023) \cite{Faure_2023_WACV}, we leverage different interaction blocks to enrich the action features of the target individual and obtain a final "interaction feature."

For the text part, we first pass each label name through the pretrained text encoder of CLIP to obtain the original text feature. Then, adapted from ~\cite{ni2022expanding}, instead of using the video feature to prompt each label, we use our interaction feature for prompting because the interaction feature can describe the action of each target person in more details. Through this approach, although the label features are fixed at the beginning, after prompting, each target person can have a set of label features of his own,  thus leading to more accurate action detection.

\subsection{Interaction Feature Generation}
Because human actions usually interact with other people or objects, we use the Person interaction block to model person-person interaction. Similarly, person-object interaction will also be considered by the Object interaction block. In addition, we use the Context interaction block to observe the target person's action from the perspective of the whole image. Finally, considering that the action contains temporal motion, we use the memory block to take into account the information before and after the current frame. Each interaction block mainly uses the attention layer to model the relationship between different features. We use the feature of the target person as the initial query, and use different features as the key and value in each block, making the target person feature contain more information step by step.

\begin{figure}[t]
\centering
   \includegraphics[width=0.4\textwidth]{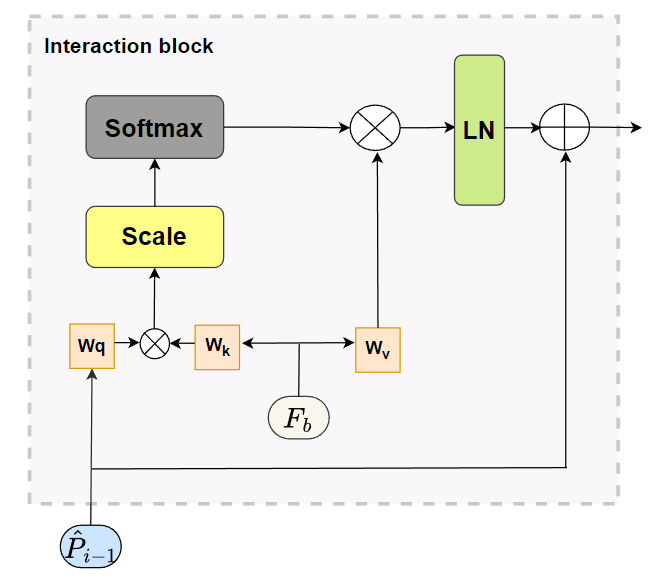}
   \caption{The interaction block takes two inputs. One is the enhanced person feature $\hat{P}_{i-1}$, and the other is the person, object, context, or memory feature $F_b$ depending on the current block.}
\label{fig:interaction unit}
\end{figure}

In more detail, the Person interaction block is a self-attention layer (SA), and its computation is given by
\begin{equation}
\begin{aligned}
&\Bar{P} = SA(P) = softmax(\frac{w_q(P) \times w_k(P)^T}{\sqrt{D}}) \times w_v(P) \\
&\hat{P} = P + LN(\Bar{P}),
\end{aligned}
\end{equation}
where ${P}$ is a batch of all the person features in the current frame, D is the channel dimension of each person feature, ${w_q}$, ${w_k}$, ${w_v}$ are used to project query, key, and value, respectively, and ${LN}$ denotes the layer normalization. As for the Object, Context, and Memory blocks, they are cross-attention layers (CA) that take two inputs. One is a batch of the "enhanced" person feature, {\ie} the output from the last interaction block, and the other input depends on the type of this block. For the object feature, we take the objects in the current frame which are detected by a pre-trained object detector. And we use the image feature of the whole current frame as our context feature. Then for the memory feature, we take the context of a certain number of neighboring frames. Figure \ref{fig:interaction unit} is the illustration of the interaction block. The following equations roughly describe how these blocks work:
\begin{equation}
\begin{aligned}
\begin{aligned}
\Bar{P}_{i-1} & = CA(\hat{P}_{i-1},F_{b}) \\
& = softmax(\frac{w_q(\hat{P}_{i-1}) \times w_k(F_{b})^T}{\sqrt{D}}) \times w_v(F_{b}) \\
\hat{P}_{i} & = \hat{P}_{i-1} + LN(\Bar{P}_{i-1})
\end{aligned}
\end{aligned}
\end{equation}
where $\hat{P}_{i-1}$ is the output of the last interaction block, and $F_b$ is the object, context, or memory feature.

In addition, since the information of the current frame makes an important role in classifying the action, after we use these interaction blocks to compute the enhanced feature of the target person, we take the mean pooling of the enhanced feature and the context feature as our final "interaction feature".

\subsection{Interaction-Aware Prompting}
For each target person, to have a set of label features that suit him/her best, we adopt an approach from prior research \cite{ni2022expanding}, which introduced a learnable prompting scheme for generating textual representations automatically. Building on this work, we develop a module that is specific to each person's interaction with one's surrounding things, which takes the interaction feature ${\hat{P}}$ to prompt with each label ${C_i}$. The module is made up of multiple blocks, each of which includes a multi-head self-attention mechanism (MSA) followed by a feed-forward network (FFN) that learns the prompts.
\begin{equation}
\begin{aligned}
    &\Bar{C} = {C} + MSA(C, \hat{P}),\\
    &\hat{C} = \Bar{C} + FFN(\bar{C}),
\end{aligned}
\end{equation}
where ${C}$ is the original text embedding, and $\hat{C}$ is the interaction-aware prompts. By treating the text embedding ${C}$ as the query and the interaction feature $\hat{P}$ as key-value pair to extract relevant interaction context from a target person. The text representation is able to extract and incorporate the relevant interaction context. Essentially, the technique allows the text embedding to capture not only the content of the text but also the relevant context in which it was produced. Subsequently, we enhance the text embedding ${C}$ by incorporating the interaction-aware prompt ${\hat{C}}$ as follows:
\begin{equation}
\begin{aligned}
    & \Tilde{C} = C + \hat{C},
\end{aligned}
\end{equation}
where ${\Tilde{C}}$ is the final textual representation to calculate similarity with interaction features. The resulting enhanced text embedding contains both the original text information and the relevant interaction context, which can improve the accuracy and effectiveness of downstream natural language processing tasks.

\subsection{Training and Inference}
During training, we take a batch of $N$ person boxes for classifying their actions with the set of training labels. For visual part, we obtain $N$ interaction feature $\hat{P}\in{\mathbb{R}^{N\times{D}}}$, where $D$ is the feature dimension. For text, we first use a text encoder to take the original text embedding of training labels, and after our Interaction-Aware Prompting, we obtain the final text embedding ${\Tilde{C}}\in{\mathbb{R}^{N\times{L}\times{D}}}$, where $L$ is the number of training labels.

For each target person in this batch, we calculate the cosine similarity between the interaction feature ${p_i}\in{\mathbb{R}^{D}}$ and the label features suitable for the person $\bar{C}\in{\mathbb{R}^{L\times{D}}}$, the training objective is to make $p_i$ and its paired action label $c_i$ have the highest similarity among $\bar{C}$. The model training is achieved by optimizing the following loss:
\begin{equation}
\begin{aligned}
    & L = -\frac{1}{N}\sum_{i=1}^{N}(log\frac{exp(p_i\cdot{c_i}/\tau)}
    {\sum_{j=1}^{L}exp(p_i\cdot{c_j}/\tau)}),
\end{aligned}
\end{equation}
where $p_i$ and $c_j$ are L2-normalized, $\tau$ is a fixed temperature parameter.
\section{Experimental Results}
\label{Experiment}
\subsection{Datasets}
\textbf{J-HMDB} dataset \cite{jhuang2013towards} has 21 action classes and 928 videos in total, which contains 31,838 annotated frames. There are up to 55 clips per class and each video clip has 15 to 40 frames, which is trimmed to contain a single action. We report frame mAP results on split-1 of the dataset. The IoU threshold for frame mAP is 0.5.

\textbf{UCF101-24} dataset \cite{soomro2012ucf101} consists of 24 action categories with 3207 untrimmed videos with human bounding boxes annotated frame by frame selected from the original UCF dataset, which contains 101 action categories. We test our method on the first split and report frame mAP with an IoU threshold of 0.5.

\subsection{Implementation Details}
\noindent{\textbf{Person and Object Detector:}} We use Faster-RCNN \cite{ren2015faster} with ResNet-50-FPN \cite{lin2017feature} backbone as our object detector, which is pretrained on ImageNet, and fine-tuned on MSCOCO. For person bounding boxes, we take groundtruth boxes for training, and use the detected boxes from \cite{kopuklu2019you} at inference time.

\vspace{1mm}

\noindent{\textbf{J-HMDB:}} We train the network for 7k iterations with 0.7k iterations serving as linear warmup and the base learning rate of 0.0002. Using SGD optimizer and 32 batch size to train the model on 4 GPUs.

\vspace{1mm}

\noindent{\textbf{UCF101-24:}} We train the network for 10k iterations with 1k iterations serving as linear warmup and the decreased base learning rate of 0.0002. We use SGD optimizer and 64 batch size to train the model on 8 GPUs.

\subsection{Zero-Shot Spatial-Temporal Action Detection}
\begin{table}[htbp]
\setlength{\tabcolsep}{4.5mm}
    \centering
    \begin{tabular}{cccc}
    \hline
    \hline
    Dataset & model &  & +IAP\\
    \hline
    \hline
    \multirow{2}*{J-HMDB} & Baseline & 59.46 & 57.41\\
     & \nickname & \textbf{65.41} & \textbf{66.83}\\
    \hline
    \hline
    \multirow{2}*{UCF101-24} & Baseline & 66.34 & 70.74\\
     & \nickname & \textbf{71.00} & \textbf{72.47} \\
    \hline
    \hline
    \end{tabular}
    \caption{
    \textbf{Zero-shot inference results
    in 75\%v.s.25\% labels split.}
     The baseline uses the image feature of whole frame for inference. +IAP: Complete model that contains Interaction-Aware Prompting.
    }
    \label{tab:Comparison_75}
\end{table}

\begin{table}[htbp]
\setlength{\tabcolsep}{4.5mm}
    \centering
    \begin{tabular}{cccc}
    \hline
    \hline
    Dataset & model &  & +IAP\\
    \hline
    \hline
    \multirow{2}*{J-HMDB} & Baseline & 42.31 & 44.55\\
     & \nickname & \textbf{44.29} & \textbf{45.18}\\
    \hline
    \hline
    \multirow{2}*{UCF101-24} & Baseline & 58.90 & \textbf{61.86}\\
     & \nickname & \textbf{59.78} & 60.30 \\
    \hline
    \hline
    \end{tabular}
    \caption{
    \textbf{Zero-shot inference results
    in 50\%v.s.50\% labels split.}
     The baseline uses the image feature of whole frame for inference. +IAP: Complete model that contains Interaction-Aware Prompting.
    }
    \label{tab:Comparison_50}
\end{table}

\begin{table}[htbp]
\setlength{\tabcolsep}{4.5mm}
    \centering
    \begin{tabular}{ccc}
    \hline
    \hline
    model &  & +prompting\\
    \hline
    \hline
    ActionCLIP \cite{wang2021actionclip} & 62.80 & 60.26\\
    Efficient-Prompting \cite{ju2022prompting} & 58.60 & 60.28 \\
     \nickname & \textbf{65.41} & \textbf{66.83}\\
    \hline
    \hline
    \end{tabular}
    \caption{
    \textbf{Comparison with CLIP-based Methods on J-HMDB.}
    The experiments are
conducted in 75\%v.s.25\% labels split on J-HMDB \cite{jhuang2013towards}.
    }
    \label{tab:jhmdb_compare}
\end{table}

In the zero-shot scenario, the training labels and testing labels are disjoint, where $C_{train} \cap C_{test} = \varnothing$. Since this scenario has not been studied in the spatial-temporal action detection before, we follow the setting proposed by \cite{ju2022prompting}. In more detail, we evaluate two settings on J-HMDB and UCF101-24: (1) taking 75\% action categories for training and the remaining 25\% for testing. (2) taking 50\% action categories for training, and the remaining 50\% for testing. Both settings use random sampling to split action categories. For the following results, we report the frame mAP with 0.5 $IoU$ threshold.

To the best of our knowledge, in this scenario, there are no prior studies with which we can compare our model. To address this issue, we propose a naive baseline based on CLIP model that does not require further training. At inference time, the baseline takes two inputs, one is the frame where the target person is located, and the other is the test label set. It then utilizes a text encoder to capture textual features and an image encoder to extract visual features from the entire image. Both encoders are obtained from the pretrained CLIP model ViT-B/16. We then measure the similarity between these two types of features as the classification scores. There are other options for the implementation of baseline, for example, taking only a person crop as a visual feature, but it will give a poor performance since it contains less information than the whole image. The only disadvantage of the baseline is that when two or more people are doing different actions in the same frame, it will classify them into the same action category. Because there is no such scenario on J-HMDB and UCF101-24, the baseline can perform well on these datasets without training.

Additionally, to conduct a fair comparison with our model, we also use the entire image features to prompt each label based on the original baseline (baseline + prompting). Table \ref{tab:Comparison_75} shows the zero-shot inference results in 75\% v.s. 25\% labels split. Without prompting, our model outperforms the baseline by 5.95 mAP on J-HMDB, and also 4.66 mAP on UCF101-24. This indicates that our interaction feature can more accurately describe the action than the naive image feature, and thus shows that our model can effectively combine different CLIP features to better recognize unseen actions. On the other hand, the zero-shot inference results are compared for 50\% v.s. 50\% labels split with the same setting as 75\% v.s. 25\% presented in Table \ref{tab:Comparison_50}. On J-HMDB, our model achieves an improvement of over 1.98 mAP without prompting and over 0.6 mAP with prompting. On UCF101-24, our model achieves an improvement of over 0.88 mAP without prompting.

\subsection{Comparison with CLIP-based Methods}

To further demonstrate the efficacy of \nickname, we experiment with other methods on our setting. Because in the J-HMDB dataset, each video only has one action label, we choose to use other zero-shot action recognition methods to classify videos into an action category, and annotate all detected person boxes in the video with this action label. For a fair comparison, we choose ActionCLIP \cite{wang2021actionclip} and Efficient-Prompting \cite{ju2022prompting} which also utilize CLIP encoders and different prompting strategies for zero-shot inference. Also, we freeze the CLIP image and text encoders for the following experiments. For both methods, they use a temporal transformer to consider the temporal relationship across frames then get the video feature. As for prompting, ActionCLIP \cite{wang2021actionclip} designs several discrete prompts which are human-readable, while Efficient-Prompting \cite{ju2022prompting} chooses to train continuous prompt vectors to search the most suitable prompts.

The result in Table \ref{tab:jhmdb_compare} suggests that our method has the best performance with or without prompting for labels. When there is no prompting, all of these methods have the 
same text embedding of each label, so the performance depends on the ability of the visual feature to describe the action. This indicates that even if the other two methods use the video feature, which considers the information of the entire video, our interaction feature can still achieve better results since it describes actions from a more granular perspective. The result also shows that, our Interaction-Aware Prompting can generate more suitable label features for each person than the hand-craft prompts used by ActionCLIP, which has worse performance after prompting.

\subsection{Ablation Study}

\begin{table*}

\vspace{2mm}
\setlength{\tabcolsep}{11mm}
    \centering
    \begin{tabular}{c|cccc|c}
    \hline
    \hline
    Dataset  & ${I_P}$ & ${I_C}$ & ${I_O}$ & ${I_M}$ & mAP \\
    \hline
    \hline
     \multirow{4}*{J-HMDB} & \checkmark & & & & 59.79 \\
     & \checkmark & \checkmark & & & 62.36 \\
     & \checkmark & \checkmark & \checkmark & & 64.82 \\
     & \checkmark & \checkmark & \checkmark & \checkmark & \textbf{65.41}  \\
    \hline
    \hline
     \multirow{4}*{UCF101-24} & \checkmark & & & & 66.92 \\
     & \checkmark & \checkmark & & & 70.34 \\
     & \checkmark & \checkmark & \checkmark & & 70.90 \\
     & \checkmark & \checkmark & \checkmark & \checkmark & \textbf{71.00}\\
    \hline
    \hline
    \end{tabular}
    \caption{\textbf{Ablation study on importance of interaction units}:
    This experiment shows the importance of various interaction units.
    ${I_P}$: Person unit; ${I_O}$: Object unit; ${I_C}$: Context unit; ${I_M}$: Memory unit. The experiments are conducted in 75\% v.s. 25\% labels split and without prompting (IAP)
    }
    \label{tab:unit importance}
\end{table*}

\begin{table*}

\vspace{2mm}
\setlength{\tabcolsep}{11mm}
    \centering
    \begin{tabular}{c|cccc|cc}
    \hline
    \hline
    Dataset  & 1st & 2nd & 3rd & 4th & mAP \\
    \hline
    \hline
    \multirow{6}*{J-HMDB} & \color{blue}${I_P}$ & \color{yellow}${I_O}$ & \color{cyan}${I_C}$ & ${I_M}$ & \textbf{65.41} \\
     & \color{blue}${I_P}$ & \color{cyan}${I_C}$ & \color{yellow}${I_O}$ & ${I_M}$ & 64.32  \\
     & \color{yellow}${I_O}$ & \color{blue}${I_P}$ & \color{cyan}${I_C}$ & ${I_M}$ & 64.68  \\
     & \color{yellow}${I_O}$ & \color{cyan}${I_C}$ & \color{blue}${I_P}$ & ${I_M}$ & 62.32 \\
     & \color{cyan}${I_C}$ & \color{blue}${I_P}$ & \color{yellow}${I_O}$ & ${I_M}$ & 62.55  \\
     & \color{cyan}${I_C}$ & \color{yellow}${I_O}$ & \color{blue}${I_P}$ & ${I_M}$ & 62.37  \\
    \hline
    \hline
    \end{tabular}
    \caption{
    \textbf{Ablation study of order of interaction units: }
    We arrange the order of units to observe the performance. In this experiment, we freeze memory units at last ordering, and the results show that the order of ${I_P-> I_O-> I_C ->I_M}$ has the best performance. The experiments are conducted in 75\% v.s. 25\% labels split and without prompting (IAP)
    }
    \label{tab:Order}
\end{table*}

\begin{table}
\vspace{2mm}
\setlength{\tabcolsep}{4.3mm}
    \centering
    \begin{tabular}{cccc|cc}
    \hline
    \hline
    ${I_P}$ & ${I_C}$ & ${I_O}$ & ${I_M}$ &  &+IAP \\
    \hline
    \hline
     \checkmark & & & &66.92 &70.74 \\
     \checkmark & \checkmark & & &70.34 &70.84 \\
     \checkmark & \checkmark & \checkmark & &70.90 &72.18 \\
     \checkmark & \checkmark & \checkmark & \checkmark & \textbf{71.00} &\textbf{72.47}  \\
    \hline
    \hline
    \end{tabular}
    \caption{\textbf{Ablation study of Interaction-Aware prompting}:
    This experiment show the effects of Interaction-Aware prompting.
    ${I_P}$: Person unit; ${I_O}$: Object unit; ${I_C}$: Context unit; ${I_M}$: Memory unit; +IAP: Complete model that contains Interaction-Aware Prompting. The experiments are conducted in 75\% v.s. 25\% labels split on UCF101-24.
    }
    \label{tab:prompting}
\end{table}

\begin{figure*}[thbp]
\begin{tabular}{llllll}

\includegraphics[width=0.25\textwidth]{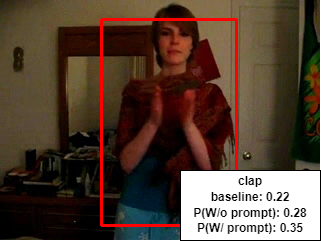}

\includegraphics[width=0.25\textwidth]{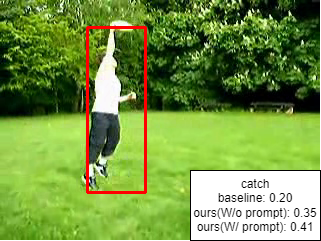}

\includegraphics[width=0.25\textwidth]{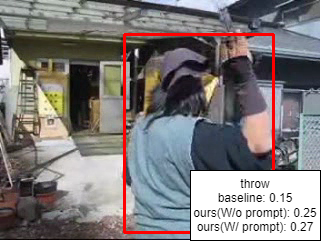}

\includegraphics[width=0.26\textwidth]{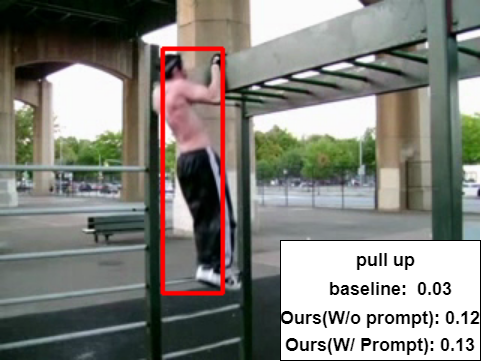}
\end{tabular}
\caption{The qualitative results on J-HMDB, which is tested on 25\% unseen labels with confidence score. The labels from left to right are ”clip", "catch", "throw", "pull up" respectively.}
\label{Fig:Race}
\end{figure*}

\begin{figure*}[thbp]
\begin{tabular}{llllll}
\includegraphics[width=0.25\textwidth]{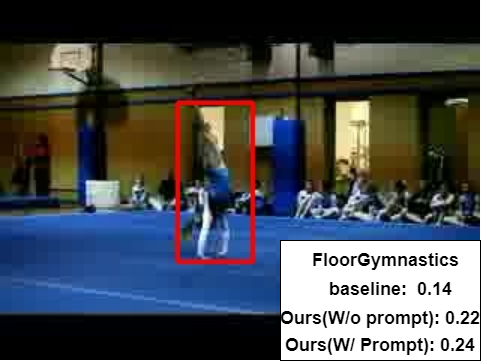}

\includegraphics[width=0.25\textwidth]{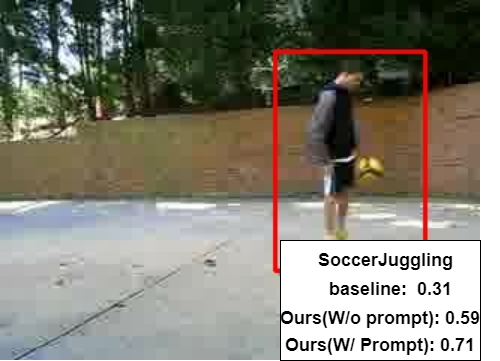}

\includegraphics[width=0.25\textwidth]{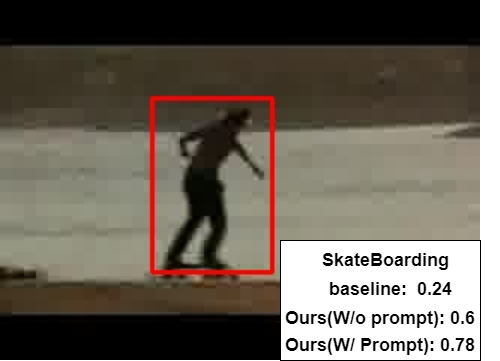}

\includegraphics[width=0.25\textwidth]{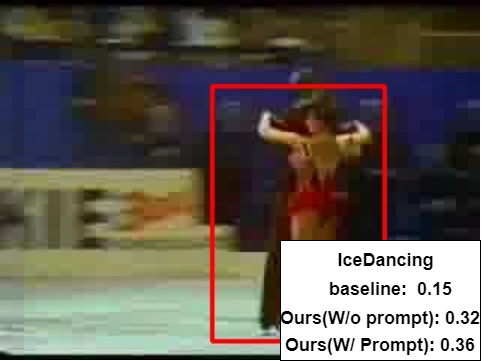}
\end{tabular}
\caption{The qualitative results on UCF101-24, which is tested on 25\% unseen labels  with confidence score. The labels from left to right are ”Floor Gymnastics", "Soccer Juggling", "Skate Boarding", "Ice Dancing" respectively.}
\label{Fig:Race_UCF}
\end{figure*}

\noindent{\textbf{Importance of interaction units :}} We conduct further investigation into the importance of different interaction units and present the results in Table \ref{tab:unit importance}. When there is no prompting, the text embedding of each label remains unchanged, so the performance only depends on whether the visual feature can be more similar to the correct label embedding. The result shows that on both J-HMDB and UCF101-24 datasets when we plug in more interaction units to aggregate different CLIP features, the performance will gradually improve. This proves that each interaction unit can indeed help to describe the action more completely, thus providing better alignment between visual and text features. With all four types of units, the model combines the most information and achieves the highest performance.

\vspace{5pt}

\noindent{\textbf{Unit order:}} In order to investigate whether the order of interaction units affects the model's performance, we conduct experiments that change the order of units so that we will not always go through the person unit first. However, we did not provide extensive experiments for this part, as previous research \cite{Faure_2023_WACV} has already explored this topic. Also, we freeze the memory unit ${I_M}$ into the last order since the memory is used to model the continuity of the action, it is more reasonable to integrate the information of the current frame before we observe the neighboring frames. The results show that changing the order of units does not lead to any improvement in performance, as indicated in the table \ref{tab:Order}. Nonetheless, we find that the best performance is achieved when the interaction blocks are ordered as ${I_P->I_O->I_C->I_M}$. The second best performance is achieved when ${I_O}$ in the first order and ${I_P}$ in the second order. We analyze that when describing an action, it is better to take into account the important local information (surrounding person/object) first, and then further use the overall scene (context) to supplement the description.

\vspace{5pt}

\noindent{\textbf{Interaction-Aware Prompting:}} In Table \ref{tab:prompting}, we conduct experiments to assess the impact of Interaction-Aware prompting on our model's performance. We find that it can help improve performance by using the interaction feature to prompt each label. The result shows that no matter in which combination, it has a higher frame mAP
 than the results without prompting in Table \ref{tab:unit importance}. In addition, in the case of prompting, the performance improves when we plug in more types of interaction units. This also indicates that our interaction feature is suitable for prompting because with more complete information, we can describe each label more appropriately.

\subsection{Visualization}
In order to better show our action detection results, we select some examples from J-HMDB and UCF101-24 and show the qualitative results with 75\%v.s.25\% labels split. In Figure \ref{Fig:Race}, for recognition of the unseen action "clap", the information of the person (e.g., pose) is very critical. We take this into account by the person interaction block of our model, and the result indeed shows that with the person block only, we already have a higher confidence score than the baseline, which only uses the whole image to infer the action. For other actions like "catch", it is necessary to consider the relationship between people and objects, as well as the continuity of actions, which can be done by our object unit and memory interaction unit, respectively. In Figure \ref{Fig:Race_UCF}, our improvement for some examples in UCF101-24 is also evident. For the examples "SoccerJuggling" and "SkateBoarding", it is difficult for the baseline to use only the whole frame to recognize, while our model can improve the confidence score to almost 0.7 and 0.8, which indicates that with our interaction module and Interaction-Aware Prompting, we can effectively aggregate more information and make visual features closer to the correct label embedding.

\label{6_Conclusion}
\section{Conclusion}

In this work, we propose a novel method for zero-shot spatio-temporal action detection, a task that has not been studied much  before. To address this problem, we introduce the \model, which leverages interaction context by using different interaction blocks to extract surrounding information. The Interaction-Aware Prompting technique is designed to generate instance-level discriminative textual representations. With these, our model can make the pre-trained visual-language model provide better alignment, thus leading to more accurate action detection for unseen action classes. Extensive experiments show that our \nickname \space surpasses the baseline under a variety of zero-shot settings. Overall, this work presents a new and effective approach for zero-shot action detection by leveraging interaction context and prompts.

\noindent{\textbf{Acknowledgements:}} This work was supported in part by the National Science and Technology Council, Taiwan under grants NSTC-111-2221-E-007-106-MY3 and NSTC-111-2634-F-007-010. We also thank National Center for High-performance Computing in Taiwan for providing  computational and storage resources. \\ \vspace{-2em}
{\small
\bibliographystyle{ieee_fullname}
\bibliography{arxiv}

\begin{thebibliography}{10}\itemsep=-1pt

\bibitem{chen2021elaborative}
Shizhe Chen and Dong Huang.
\newblock Elaborative rehearsal for zero-shot action recognition.
\newblock In {\em Proceedings of the IEEE/CVF International Conference on
  Computer Vision}, pages 13638--13647, 2021.

\bibitem{Faure_2023_WACV}
Gueter~Josmy Faure, Min-Hung Chen, and Shang-Hong Lai.
\newblock Holistic interaction transformer network for action detection.
\newblock In {\em Proceedings of the IEEE/CVF Winter Conference on Applications
  of Computer Vision (WACV)}, pages 3340--3350, January 2023.

\bibitem{feichtenhofer2019slowfast}
Christoph Feichtenhofer, Haoqi Fan, Jitendra Malik, and Kaiming He.
\newblock Slowfast networks for video recognition.
\newblock In {\em Proceedings of the IEEE/CVF international conference on
  computer vision}, pages 6202--6211, 2019.

\bibitem{gan2015exploring}
Chuang Gan, Ming Lin, Yi Yang, Yueting Zhuang, and Alexander~G Hauptmann.
\newblock Exploring semantic inter-class relationships (sir) for zero-shot
  action recognition.
\newblock In {\em Proceedings of the AAAI Conference on Artificial
  Intelligence}, 2015.

\bibitem{girdhar2019video}
Rohit Girdhar, Joao Carreira, Carl Doersch, and Andrew Zisserman.
\newblock Video action transformer network.
\newblock In {\em Proceedings of the IEEE/CVF Conference on Computer Vision and
  Pattern Recognition}, pages 244--253, 2019.

\bibitem{Jain_2015_ICCV}
Mihir Jain, Jan~C. van Gemert, Thomas Mensink, and Cees G.~M. Snoek.
\newblock Objects2action: Classifying and localizing actions without any video
  example.
\newblock In {\em Proceedings of the IEEE International Conference on Computer
  Vision (ICCV)}, December 2015.

\bibitem{jhuang2013towards}
Hueihan Jhuang, Juergen Gall, Silvia Zuffi, Cordelia Schmid, and Michael~J
  Black.
\newblock Towards understanding action recognition.
\newblock In {\em Proceedings of the IEEE international conference on computer
  vision}, pages 3192--3199, 2013.

\bibitem{jia2021scaling}
Chao Jia, Yinfei Yang, Ye Xia, Yi-Ting Chen, Zarana Parekh, Hieu Pham, Quoc Le,
  Yun-Hsuan Sung, Zhen Li, and Tom Duerig.
\newblock Scaling up visual and vision-language representation learning with
  noisy text supervision.
\newblock In {\em International Conference on Machine Learning}, pages
  4904--4916. PMLR, 2021.

\bibitem{ju2022prompting}
Chen Ju, Tengda Han, Kunhao Zheng, Ya Zhang, and Weidi Xie.
\newblock Prompting visual-language models for efficient video understanding.
\newblock In {\em Computer Vision--ECCV 2022: 17th European Conference, Tel
  Aviv, Israel, October 23--27, 2022, Proceedings, Part XXXV}, pages 105--124.
  Springer, 2022.

\bibitem{kopuklu2019you}
Okan K{\"o}p{\"u}kl{\"u}, Xiangyu Wei, and Gerhard Rigoll.
\newblock You only watch once: A unified cnn architecture for real-time
  spatiotemporal action localization.
\newblock {\em arXiv preprint arXiv:1911.06644}, 2019.

\bibitem{li2019collaborative}
Chao Li, Qiaoyong Zhong, Di Xie, and Shiliang Pu.
\newblock Collaborative spatiotemporal feature learning for video action
  recognition.
\newblock In {\em Proceedings of the IEEE/CVF Conference on Computer Vision and
  Pattern Recognition}, pages 7872--7881, 2019.

\bibitem{lin2017single}
Tianwei Lin, Xu Zhao, and Zheng Shou.
\newblock Single shot temporal action detection.
\newblock In {\em Proceedings of the 25th ACM international conference on
  Multimedia}, pages 988--996, 2017.

\bibitem{lin2017feature}
Tsung-Yi Lin, Piotr Doll{\'a}r, Ross Girshick, Kaiming He, Bharath Hariharan,
  and Serge Belongie.
\newblock Feature pyramid networks for object detection.
\newblock In {\em Proceedings of the IEEE conference on computer vision and
  pattern recognition}, pages 2117--2125, 2017.

\bibitem{mandal2019out}
Devraj Mandal, Sanath Narayan, Sai~Kumar Dwivedi, Vikram Gupta, Shuaib Ahmed,
  Fahad~Shahbaz Khan, and Ling Shao.
\newblock Out-of-distribution detection for generalized zero-shot action
  recognition.
\newblock In {\em Proceedings of the IEEE/CVF Conference on Computer Vision and
  Pattern Recognition}, pages 9985--9993, 2019.

\bibitem{materzynska2020something}
Joanna Materzynska, Tete Xiao, Roei Herzig, Huijuan Xu, Xiaolong Wang, and
  Trevor Darrell.
\newblock Something-else: Compositional action recognition with
  spatial-temporal interaction networks.
\newblock In {\em Proceedings of the IEEE/CVF Conference on Computer Vision and
  Pattern Recognition}, pages 1049--1059, 2020.

\bibitem{mettes2022universal}
Pascal Mettes.
\newblock Universal prototype transport for zero-shot action recognition and
  localization.
\newblock {\em arXiv preprint arXiv:2203.03971}, 2022.

\bibitem{mettes2017spatial}
Pascal Mettes and Cees~GM Snoek.
\newblock Spatial-aware object embeddings for zero-shot localization and
  classification of actions.
\newblock In {\em Proceedings of the IEEE international conference on computer
  vision}, pages 4443--4452, 2017.

\bibitem{Mettes_2017_ICCV}
Pascal Mettes and Cees G.~M. Snoek.
\newblock Spatial-aware object embeddings for zero-shot localization and
  classification of actions.
\newblock In {\em Proceedings of the IEEE International Conference on Computer
  Vision (ICCV)}, Oct 2017.

\bibitem{mettes2021object}
Pascal Mettes, William Thong, and Cees~GM Snoek.
\newblock Object priors for classifying and localizing unseen actions.
\newblock {\em International Journal of Computer Vision}, 129:1954--1971, 2021.

\bibitem{nag2022zero}
Sauradip Nag, Xiatian Zhu, Yi-Zhe Song, and Tao Xiang.
\newblock Zero-shot temporal action detection via vision-language prompting.
\newblock In {\em Computer Vision--ECCV 2022: 17th European Conference, Tel
  Aviv, Israel, October 23--27, 2022, Proceedings, Part III}, pages 681--697.
  Springer, 2022.

\bibitem{ni2022expanding}
Bolin Ni, Houwen Peng, Minghao Chen, Songyang Zhang, Gaofeng Meng, Jianlong Fu,
  Shiming Xiang, and Haibin Ling.
\newblock Expanding language-image pretrained models for general video
  recognition.
\newblock In {\em Computer Vision--ECCV 2022: 17th European Conference, Tel
  Aviv, Israel, October 23--27, 2022, Proceedings, Part IV}, pages 1--18.
  Springer, 2022.

\bibitem{pan2021actor}
Junting Pan, Siyu Chen, Mike~Zheng Shou, Yu Liu, Jing Shao, and Hongsheng Li.
\newblock Actor-context-actor relation network for spatio-temporal action
  localization.
\newblock In {\em Proceedings of the IEEE/CVF Conference on Computer Vision and
  Pattern Recognition}, pages 464--474, 2021.

\bibitem{Pan_2021_CVPR}
Junting Pan, Siyu Chen, Mike~Zheng Shou, Yu Liu, Jing Shao, and Hongsheng Li.
\newblock Actor-context-actor relation network for spatio-temporal action
  localization.
\newblock In {\em Proceedings of the IEEE/CVF Conference on Computer Vision and
  Pattern Recognition (CVPR)}, pages 464--474, June 2021.

\bibitem{qin2017zero}
Jie Qin, Li Liu, Ling Shao, Fumin Shen, Bingbing Ni, Jiaxin Chen, and Yunhong
  Wang.
\newblock Zero-shot action recognition with error-correcting output codes.
\newblock In {\em Proceedings of the IEEE Conference on Computer Vision and
  Pattern Recognition}, pages 2833--2842, 2017.

\bibitem{radford2021learning}
Alec Radford, Jong~Wook Kim, Chris Hallacy, Aditya Ramesh, Gabriel Goh,
  Sandhini Agarwal, Girish Sastry, Amanda Askell, Pamela Mishkin, Jack Clark,
  et~al.
\newblock Learning transferable visual models from natural language
  supervision.
\newblock In {\em International conference on machine learning}, pages
  8748--8763. PMLR, 2021.

\bibitem{ren2015faster}
Shaoqing Ren, Kaiming He, Ross Girshick, and Jian Sun.
\newblock Faster r-cnn: Towards real-time object detection with region proposal
  networks.
\newblock {\em Advances in neural information processing systems}, 28, 2015.

\bibitem{soomro2012ucf101}
Khurram Soomro, Amir~Roshan Zamir, and Mubarak Shah.
\newblock Ucf101: A dataset of 101 human actions classes from videos in the
  wild.
\newblock {\em arXiv preprint arXiv:1212.0402}, 2012.

\bibitem{tang2020asynchronous}
Jiajun Tang, Jin Xia, Xinzhi Mu, Bo Pang, and Cewu Lu.
\newblock Asynchronous interaction aggregation for action detection.
\newblock In {\em Proceedings of the European conference on computer vision
  (ECCV)}, 2020.

\bibitem{tong2022videomae}
Zhan Tong, Yibing Song, Jue Wang, and Limin Wang.
\newblock Videomae: Masked autoencoders are data-efficient learners for
  self-supervised video pre-training.
\newblock {\em Advances in neural information processing systems},
  35:10078--10093, 2022.

\bibitem{wang2021actionclip}
Mengmeng Wang, Jiazheng Xing, and Yong Liu.
\newblock Actionclip: A new paradigm for video action recognition.
\newblock {\em arXiv preprint arXiv:2109.08472}, 2021.

\bibitem{wu2019long}
Chao-Yuan Wu, Christoph Feichtenhofer, Haoqi Fan, Kaiming He, Philipp
  Krahenbuhl, and Ross Girshick.
\newblock Long-term feature banks for detailed video understanding.
\newblock In {\em Proceedings of the IEEE/CVF Conference on Computer Vision and
  Pattern Recognition}, pages 284--293, 2019.

\bibitem{yang2021beyond}
Xitong Yang, Haoqi Fan, Lorenzo Torresani, Larry~S Davis, and Heng Wang.
\newblock Beyond short clips: End-to-end video-level learning with
  collaborative memories.
\newblock In {\em Proceedings of the IEEE/CVF Conference on Computer Vision and
  Pattern Recognition}, pages 7567--7576, 2021.

\bibitem{yuan2021florence}
Lu Yuan, Dongdong Chen, Yi-Ling Chen, Noel Codella, Xiyang Dai, Jianfeng Gao,
  Houdong Hu, Xuedong Huang, Boxin Li, Chunyuan Li, et~al.
\newblock Florence: A new foundation model for computer vision.
\newblock {\em arXiv preprint arXiv:2111.11432}, 2021.

\bibitem{zhou2021graph}
Jiaming Zhou, Kun-Yu Lin, Haoxin Li, and Wei-Shi Zheng.
\newblock Graph-based high-order relation modeling for long-term action
  recognition.
\newblock In {\em Proceedings of the IEEE/CVF Conference on Computer Vision and
  Pattern Recognition}, pages 8984--8993, 2021.

\end{thebibliography}
}
\clearpage
\appendix
\renewcommand\thesection{\Alph{section}}
\begin{table*}
\vspace{-40em}
\setlength{\tabcolsep}{6.3mm}
    \centering
    \begin{tabular}{c|ccccccc}
    \hline
    \hline
     model & catch & clap & pullup & sit & throw & wave & mAP\\
    \hline
    \hline
    Baseline & 69.70 & 45.95 & \textbf{99.98} & \textbf{41.87} & 33.57 & \textbf{65.71} &  59.46\\
    \nickname & \textbf{81.66} & \textbf{66.74} & 99.94 & 38.46 & \textbf{52.22} & 53.42 & \textbf{65.41} \\
    \hline
    \hline
    \end{tabular}
    \caption{
    \textbf{Frame AP of J-HMDB per unseen class
    in 75\%v.s.25\% labels split.}
     The baseline uses the image feature of whole frame for inference. Both baseline and \nickname \space are without prompting.
    }  
    \label{tab:JHMDB_class_compare}
\end{table*}

\begin{table*}
\vspace{-80em}
\setlength{\tabcolsep}{2mm}
    \centering
    \begin{tabular}{c|ccccccc}
    \hline
    \hline
    model & FloorGymnastics & IceDancing & SalsaSpin & SkateBoarding & SoccerJuggling & VolleyballSpiking & mAP\\
    \hline
    \hline
     Baseline & 74.69 & 65.98 & \textbf{63.92} & 91.25 & \textbf{98.64} & 3.55 & 66.34\\
    \nickname & \textbf{87.29} & \textbf{67.26} & 58.79 & \textbf{92.68} & 98.62 & \textbf{21.37} & \textbf{71.00}\\
    \hline
    \hline
    \end{tabular}
    \caption{
    \textbf{Frame AP of UCF101-24 per unseen class
    in 75\%v.s.25\% labels split.}
     The baseline uses the image feature of whole frame for inference. Both baseline and \nickname \space are without prompting.
}
    \label{tab:UCF_class_compare}
\end{table*}

\section*{Supplementary Material}
In this supplementary material, we aim to provide further analysis on our experiments, and show more experimental results for comparison with other methods.
\section{The efficacy of baseline}
To further showcase the CLIP baseline's capability in recognizing unseen actions, we directly utilize the baseline to infer all labels of the J-HMDB dataset. We compare with the methods \cite{Mettes_2017_ICCV,mettes2021object} that have done the same experiment on J-HMDB. To hava a fair comparison, we use the same perosn detector as theirs(i.e. Faster R-CNN, pre-trained on MS-COCO), and use video mAP for evaluation as they also provided. Among the settings of IoU threshold from 0.1 to 0.5, our CLIP baseline all register significant gains compared to them. When the threshold is set to 0.1, we can achieve 56.82 mAP score, while these two methods obtain 27.5 and 32.1, respectively. Note that the influence of localization error can be almost ignored when the threshold is very small, thus it can be more focused in the comparison of classification.

\section{Analysis on 50\% vs 50\% experiment}
In our study, we perform a zero-shot experiment in 50\% vs 50\% labels split and compare the performance of our model with the baseline model on the UCF101-24 dataset \cite{soomro2012ucf101}. The results in Table \ref{tab:Comparison_50} show that with prompting, the baseline model achieved a higher mAP score than our model on UCF101-24. Further analysis revealed that due to the lower resolution of UCF101-24 videos, it will lead to noisy results in the process of generating interaction features. Moreover, the 50\% vs 50\% labels split reduces the training data, which may amplify the noise and favor the baseline model that relies solely on image features. Nevertheless, the prompting mechanism enhances the performance of both our model and the baseline. These findings suggest that prompts are beneficial for this task. Additionally, our model outperforms the baseline for the 75\% vs 25\% experiment on UCF101-24, indicating that  with an appropriate amount of training data, we can still generate representative interaction features for detecting unseen actions even if the video has lower resolution.

\section{Average precision (AP) of each unseen class}
For a more detailed comparison of the results, we present the average precision (AP) of each unseen class. Table \ref{tab:JHMDB_class_compare} presents the result on J-HMDB, our model performs better on half of the classes. In addition, in these worse classes, we are only 12\% lower than the baseline at most, while the others are almost the same. On the other hand, we have made great progress in better classes, with a minimum improvement of 12\% and a maximum of almost 21\%. From Table \ref{tab:UCF_class_compare}, we can see that on UCF101-24, our model has progressed in most classes, especially for the challenging class where the baseline has only 3.55\% AP.

\section{More details of bounding box}
 For person boxes, we take groundtruth boxes at training time, and we use the boxes detected from \cite{kopuklu2019you} at inference time, which is a single-stage framework for action localization and classification. Besides, in order to avoid wrongly detected person boxes from causing noise in the interaction module, we only take boxes whose confidence scores are greater than 0.2 at inference time. Regarding object detection, we employ Faster-RCNN to detect object boxes during both training and inference. We select objects that intersect with any person (i.e. IoU $>$ 0) to capture relevant contextual information.

For the action detection framework, since we can only use part of the training data in the zero-shot setting, it is more challenging to train a localization network from scratch. Instead, we use extra person detector for localization and let our model focus on recognizing unseen actions. Notably, even in fully-supervised settings, several SOTA  methods \cite{tang2020asynchronous,Pan_2021_CVPR,tong2022videomae} exploit human detector and only focus on classification.

\section{Capability for full supervision}
For fully-supervised setting, our approach achieves frame mAP of 73.70 on J-HMDB and 78.19 on UCF101-24 respectively. Note that the CLIP encoders are frozen during training, which indicates that our tunable parameters(11.6M) are much less than other SOTA methods. In addition, we also conduct the experiment with our baseline, which obtains 48.70 on J-HMDB and 58.08 mAP on UCF101-24. No matter in fully-supervised or zero-shot setting, our method can make more effective use of visual-language features to improve performance.

\section{Advantage}
Our model has a significant advantage over other models in its class, as it requires only \textbf{11.6M} parameters for training. Compared to other models, which usually require larger numbers of parameters in the training, our model can achieve high performance without incurring as much computational cost. In particular, this advantage is especially pronounced for zero-shot scenarios, where models must be able to learn quickly and adapt to unseen data. By utilizing fewer parameters, our model is able to learn faster and more efficiently, enabling it to outperform other models for zero-shot action detection tasks.

\end{document}